\documentclass[conference]{IEEEtran}
\IEEEoverridecommandlockouts
\usepackage{cite}
\usepackage{amsmath,amssymb,amsfonts}
\usepackage{algorithmic}
\usepackage{graphicx}
\usepackage{textcomp}
\usepackage{xcolor}
\usepackage{gensymb}
\usepackage{mathpazo}
\usepackage{algorithm}
\usepackage{esvect}
\usepackage{amsmath}
\usepackage{tikz}
\usepackage{subfig}
\usepackage{xcolor}
\usepackage{todonotes}

\graphicspath{{Figures/}}

\def\BibTeX{{\rm B\kern-.05em{\sc i\kern-.025em b}\kern-.08em
    T\kern-.1667em\lower.7ex\hbox{E}\kern-.125emX}}

\begin{document}

\newcommand\pd[1]{{\color{blue}{#1}}}
\newcommand\pdmargin[1]{\marginpar{\color{orange}\tiny\ttfamily From PD: #1}}

\title{Raci-Net: Ego-vehicle Odometry Estimation in Adverse Weather Conditions}

\author{Mohammadhossein Talebi$^{1}$, Pragyan Dahal$^{2}$, Davide Possenti$^{1}$, Stefano Arrigoni$^{1}$, Francesco Braghin$^{1}$   
\thanks{$^{1}$ Politecnico di Milano, 20156, Italy}
\thanks{$^{2}$ Michigan State University, East Lansing, MI, USA}
}

\maketitle
\thispagestyle{plain}
\pagestyle{plain}
 
\begin{abstract}
Autonomous driving systems are highly dependent on sensors such as camera, LiDAR, and inertial measurement unit (IMU) to perceive the environment and estimate their motion. Among these sensors, perception-based sensors are not protected from harsh weather and technical failures. Although existing methods show robustness against common technical issues like rotational misalignment and disconnection, they often degrade when faced with dynamic environmental factors like weather conditions. To address these problems, this research introduces a novel deep learning-based motion estimator that integrates visual, inertial, and millimeter wave radar data, utilizing each sensor's strengths to improve odometry estimation accuracy and reliability under adverse environmental conditions such as snow, rain, and varying light. The proposed model uses advanced sensor fusion techniques that dynamically adjust the contributions of each sensor based on the current environmental condition, with radar compensating for visual sensor limitations in poor visibility. This work explores recent advancements in radar-based odometry and highlights that radar robustness in different weather conditions makes it a valuable component for pose estimation systems, specifically when visual sensors are degraded. Experimental results, conducted on the Boreas dataset, showcases the robustness and effectiveness of the model in both clear and degraded environments. (\textit{Code will be open-sourced upon the paper acceptance})\\
\end{abstract}


\section{Introduction}
\label{sec:Introduction}
Ego vehicle state estimation is a fundamental component of autonomous vehicle systems, responsible for determining the vehicle's current position, orientation, and motion status in real-time \cite{dahal2023vehicle,RobustStateNet,msf,Integrated_Ego_Estimation}. Odometry estimation, a subset of state estimation, involves predicting the change in pose of a mobile robot over a given time step. This process is focused on local predictions, meaning it only considers the movement within a specific time frame. Whether considering local or global estimates, accuracy is essential for maintaining a reliable understanding of the vehicle’s movement and trajectory. This accuracy is crucial in autonomous systems, as any error in motion estimation could compromise the performance of downstream operations.\par
Invariably, the overall performance of an AV system is greatly enhanced through the integration of multiple types of sensors, a method known as sensor fusion, since each sensor offers unique strengths and weaknesses. Cameras, while highly effective in clear weather, often struggle under adverse environmental conditions such as rain, snow, and low-light condition. Radars face significant challenges when operating in isolation. Due to phenomena such as specular (mirror-like) reflections, diffraction \cite{Millimeter_Wave_Human_Blockage}, and substantial multi-path effects, radar returns are often corrupted by noise. Additionally, the point cloud generated from radar scans tends to be highly sparse \cite{lu2020milliego}. These challenges have driven research to push the boundaries by integrating different sensor modalities. While prior research has focused on technical failures in sensor fusion systems, such as hardware malfunctions or signal interference, relatively little attention has been paid to failures specifically caused by adverse weather conditions. These limitations can compromise the safety and reliability of camera-based navigation systems, which are essential for localization in AVs. Recognizing these challenges, this work places a particular emphasis on the use of millimeterwave (mmWave) radar to complement visual data.\par
Radar sensors can operate in varied weather conditions due to a wide spectrum wavelength and absence of mechanical moving parts, as opposed to passive visual sensors due to their lack of scope for climate range \cite{vargas2021overview}. We introduce a novel deep learning-based motion estimator that integrates data from three distinct modalities: inertial, visual, and mmWave radar. This multi-modal approach leverages the robustness of radar to compensate for the weaknesses of cameras in adverse conditions. The radar provides reliable obstacle detection and range measurements even in situations where cameras fail, such as during precipitation or low-light condition. Meanwhile, IMUs offer precise information on the vehicle’s motion and orientation. In this work, we focus on odometry estimation, specifically the model's capability to estimate ego-motion. This means the model operates within a local coordinate system, which is redefined at the end of each previous time step for the subsequent estimation. Consequently, the estimation process functions in an open-loop configuration. The main contributions of this work are:
\begin{itemize}
    \item We propose a deep neural network architecture for multi-modal sensor fusion that leverages visual, inertial, and mmWave radar modalities for odometry estimation. 
    \item Starting from the foundation of an existing model known as "Under The Radar", a novel radar encoder was introduced to effectively capture and transform robust features from mmWave radar data, improving localization in challenging weather conditions.
    \item Comprehensive evaluations were conducted using the Boreas dataset, showcasing the model's robustness across various segment lengths and environmental conditions, leading to improvements in odometry estimation.
\end{itemize}\par
The remainder of the article is organized as follows. Section~\ref{sec:related_works} presents a survey of works in the literature related to egomotion estimation. In Section~\ref{sec:Problem}, we explain the problem statement the paper is trying to solve. Section~\ref{subsec:enconding} details the feature extraction procedure for each modality, \ref{subsec:fusion} explains the fusion strategy, and \ref{subsec:temp_pose} clarifies the remaining parts of the proposed model. Section~\ref{subsec:training_details} provides the details regarding the training of the proposed models. Then, we provide the results in Section~\ref{subsec:results}, and discuss about them in Section~\ref{sec:discussion}. Finally, Section~\ref{sec:Conclusion} concludes the paper.

\section{Related Works}
\label{sec:related_works}
\subsection{Odometry Problem}
\label{subsec:odomtery}
Traditionally, odometry has been primarily performed using wheel encoders, which was prone to errors caused by slip, surface irregularities, and mechanical issues, leading to inaccuracies over time. To address these challenges, researchers have increasingly turned to Visual Odometry (VO). Monocular and stereoscopic VO simultaneous localization and mapping (SLAM) systems, such as ORB-SLAM \cite{ORB_SLAM} and LSD-SLAM \cite{LSD_SLAM}, have shown advances in tracking motion through image sequences but face limitations in dynamic or poorly lit environments \cite{lim2020review, cheng2022review}. Inertial Odometry (IO) provides another alternative. However, like wheel encoders, IO suffers from drift over time. To further enhance accuracy, many researchers have focused on Visual-Inertial Odometry (VIO). The IMU provides acceleration and rotation data, which complements the camera's visual information by helping to reduce drift over time, and successful implementations like Project Tango have demonstrated its robustness \cite{bala2022advances}.\par
\subsection{Traditional and Deep Learning-Based Sensor Fusion}
\label{subsec:sensor_fusion_general}
Traditional sensor fusion techniques depend heavily on manually crafted designs that rely significantly on human expertise and specialized knowledge. Deep learning-based sensor fusion has emerged as a solution, offering automatic feature extraction and the ability to process multi-modal data, making it suitable for complex environments \cite{tang2023comparative}. Prominent deep learning models, such as DeepVO \cite{wang2017deepvo} and VINet \cite{clark2017vinet}, showcase how learned representations can bypass the need for manual feature design and achieve improved odometry estimation.\par
While recent advances in deep learning-based sensor fusion have significantly improved the accuracy of tasks such as localization, odometry, and motion estimation, there remain several key challenges that need to be addressed to ensure robustness in real-world applications. One of the major gaps is the lack of attention to sensor failures and environmental uncertainties. Much of the focus has been on improving the accuracy of odometry estimation, with limited work addressing how sensor fusion models handle failure scenarios. In many works, including \cite{kendall2015posenet, brahmbhatt2018geometry, clark2017vidloc}, features from different sensor modalities are fed directly into subsequent pose regression models or simply concatenated. Recent works such as \cite{chen2019selective, dahal2023fault} tried to address this by incorporating attention mechanisms that dynamically adjust the fusion based on the reliability of the input sensors. These approaches are designed to account for IMU and camera failures.\par
Another critical gap is environmental robustness. Sensor fusion models, particularly VIO systems, are known to degrade significantly under adverse environmental conditions. For instance, RGB cameras perform poorly in low-light conditions, and depth cameras struggle with glare and intense illumination, making them unreliable in certain scenarios \cite{saputra2020deeptio}. Similarly, LiDAR-Inertial Odometry (LIO) systems are susceptible to performance degradation in environments with airborne obscurants like dust, fog, or smoke, which affect the laser beams used by LiDAR sensors \cite{richardson2011strengths}. These challenges highlight the need for sensor fusion strategies that are resilient to visibility issues and adverse weather conditions.\par

\subsection{Radar-based Odometry}
\label{subsec: radar_based_odometry}
A key advantage of radar over vision sensors is in its robustness to environmental conditions. Compared with LiDAR, radars use electronic beamforming and are therefore lightweight and able to fit the payloads of micro robots and form factors of mobile or wearable devices. Furthermore, as it is Radio Frequency (RF) based, it does not require optical lenses and can be integrated into plastic housings \cite{lu2020milliego}, making them highly resilient to water and dust ingress. We therefore envision that odometry based on a radar will allow robust ego-motion estimation in complex environments and harsh weather conditions, as well as serve as a new enabler for ubiquitous mobility with mobile devices.\par
Most of the previous approaches to radar odometry have relied on handcrafted feature extraction, like \cite{cen2018precise, kung2023ndt, burnett2021motion}. Barnes and Posner \cite{barnes2020under} previously showed that learned features have the potential to outperform hand-crafted features. The authors propose a novel approach that embeds a differentiable point-based motion estimator within a convolutional neural network (CNN) architecture, allowing the network to learn keypoint locations, scores, and descriptors directly from localization errors. This method eliminates the need for manually designed keypoints. \par
Despite the growing interest in radar-based odometry, most research has primarily focused on radar-only odometry. While radar provides significant advantages in terms of robustness against environmental factors such as fog, rain, and low visibility, the limitations of using radar alone include sparse data representations, noise, and difficulty in distinguishing between static and dynamic objects. This makes the fusion of radar with complementary sensors, such as cameras or IMUs, an attractive solution. To the best of our knowledge, only a few studies explored the fusion of radar with other modalities like visual and inertial data. Lu et. al \cite{lu2020milliego} focused on fusing radar data with other modalities introducing a novel deep learning-based system named “milliEgo” for ego-motion estimation that leverages the robustness of mmWave radar combined with IMUs. Their approach addresses the challenges posed by the sparse and noisy data from mmWave radar by employing a deep neural network that directly learns motion transformations, bypassing traditional point cloud registration methods.  However, while this system adeptly integrates visual and inertial modalities with radar, it may still face challenges in ensuring the robustness and generalization of sensor fusion under diverse and adverse weather conditions that is critical for real-world autonomous navigation.\par

\section{Problem Formulation}
\label{sec:Problem}
This work focuses on the general problem of multi-modal odometry, which involves estimating a vehicle's ego-motion using raw data from various sensor types. Formally, an arbitrary sensor input can be represented as $X_{\xi} = \{ x_i \}_{i=1}^K$ where $K$ denotes the number of data points. In the case of our proposed model, we leverage a collection of sensor data $X = \{ X_M, X_I, X_V \}$ where the inputs are collected from a mmWave radar (M), an inertial sensor (I), and a visual sensor (V). The system goal is to estimate the 6-DoF ego-motion $y = [t, r]$ where $t$ represents the relative translation, and $r$ represents the relative rotation between two consecutive frames.\par
The problem is framed as a learning task: to find an effective mapping $f(x)$ that minimizes the error between the predicted motion and the ground truth motion. Specifically, the output of the odometry model is the 6-DoF relative pose, computed between pairs of consecutive sensor frames. To achieve global odometry estimation, a Special Euclidean group in 3 dimensions,$SE(3)$, composer is used to iteratively combine these relative poses, creating a 6-DoF trajectory that represents the vehicle's motion over time.\par

\section{Methodology}
\label{sec:methodology}
\begin{figure}[htbp]
    \centering
    \includegraphics[width=0.5\textwidth]{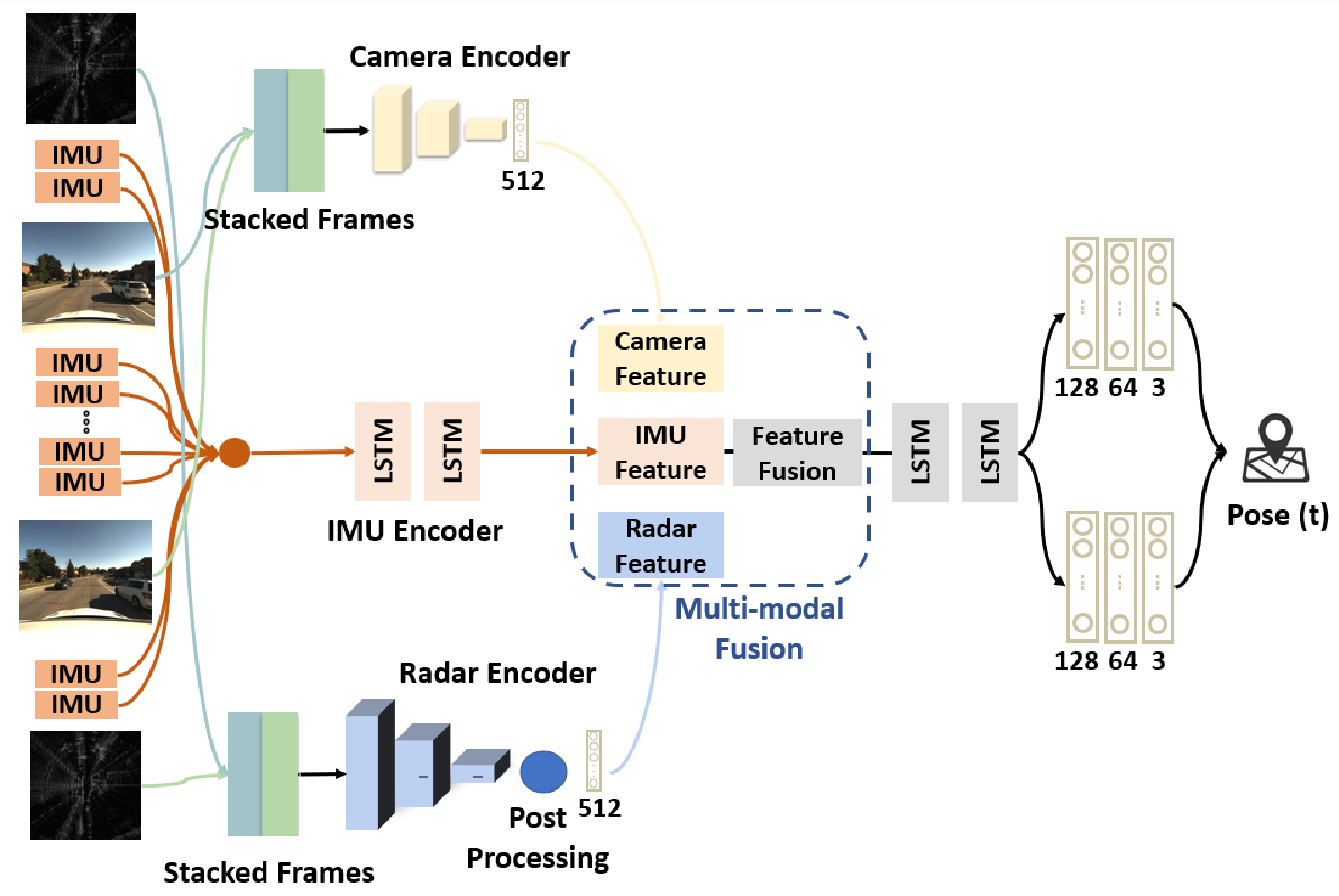}
    \caption{The proposed model architecture for multi-modal ego-motion estimation.}
    \label{fig:model_architecture}
\end{figure}

The proposed model, named RaCI-Net, shown in Fig.~\ref{fig:model_architecture}, is a multi-modal system that integrates mmWave radar, inertial, and visual data to estimate the ego-motion of a vehicle. At its core, it leverages these diverse sensor inputs within an end-to-end trainable deep learning framework designed to accurately predict 6-DoF vehicle motion. The standout feature of our model is its novel mmWave radar feature extractor subnet, which enhances the model’s capability to learn motion using highly effective radar features. These features are particularly powerful because they are extracted based on upstream, physically meaningful properties, such as the radar’s translation and rotation. This ensures that the extracted features are directly tied to the real-world spatial dynamics of the vehicle and environment, allowing the model to better estimate motion even in complex or noisy conditions.

\subsection{Feature Extraction}
\label{subsec:enconding}
\begin{figure}[H]
    \centering
    \includegraphics[width=0.3\textwidth]{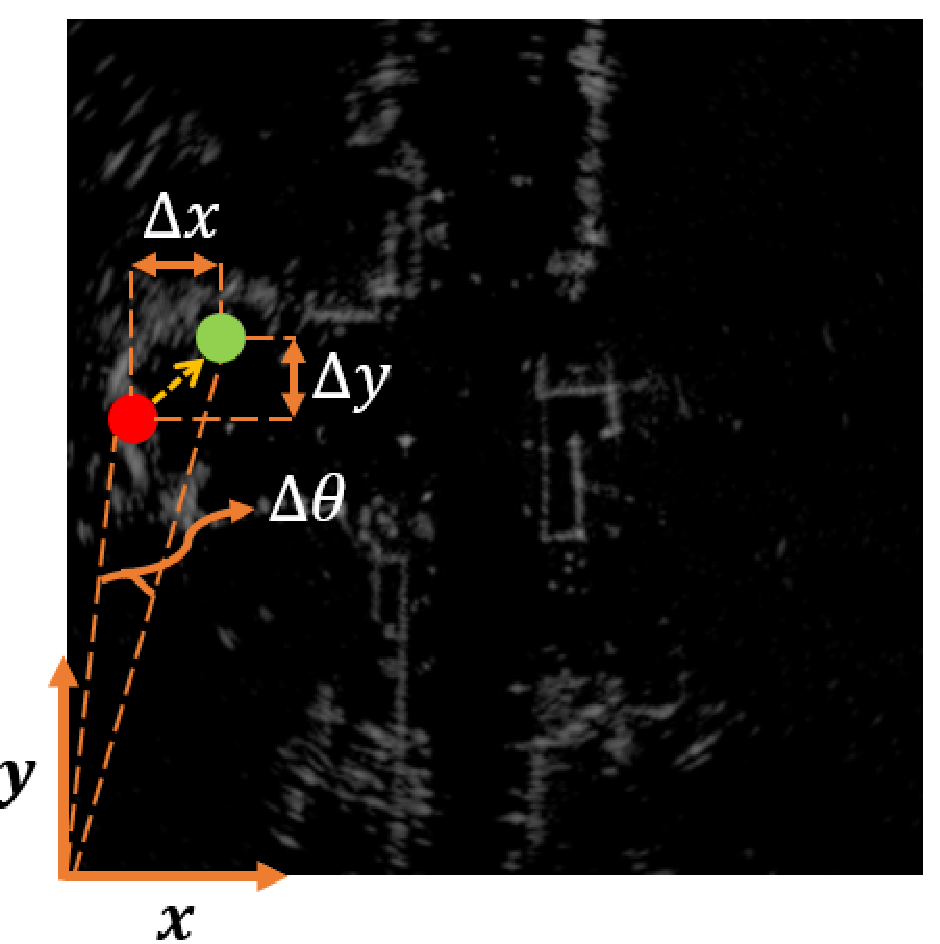}
    \caption{Visualization of post-computations between consecutive radar frames. The initial position (red point) in the first frame represents the same keypoint as the new position (green point) in the second frame, which has shifted due to movement, resulting in a computed translation $\Delta x$, $\Delta y$, and rotation $\Delta \theta$.}
    \label{fig:radar_encoder_keypoints}
\end{figure}

The model consists of dedicated encoder blocks for each sensor type—radar, camera, and IMU—to extract meaningful latent features from raw data. The camera encoder processes consecutive image frames to capture visual context and motion. It uses a modified RAFT network \cite{teed2020raft} to extract optical flow and feature representations, which are refined by adding a single fully connected (FC) layer at the end to capture high-level deep features. The radar encoder inherits the architecture outlined in the "Under The Radar" model \cite{barnes2020under}, extracting keypoints locations and descriptors from radar data and using a softmax matcher to match the keypoints in consecutive frames. Finally, we propose a post-processing unit to transform these outputs into high-level deep features that include motion information. The post-processor takes the set of keypoints locations and descriptors from both consecutive radar frames, and computes the changes in these values in this time period. It forms a 2D matrix shown in Eq.~\ref{eq:radar_2D_matrix}, in which every row is associated to a keypoint, and the columns represent the change in descriptors, translation of the keypoint ($\Delta x_{i}$ and $\Delta y_{i}$), and the translation vector angle in the plane of movement. The final matrix, which includes motion information between two frames, is flattened and enters a single FC layer in order to be transformed into deep features and reduce the dimensionality. Fig.~\ref{fig:radar_encoder_keypoints} is provided to visualize this computation.\par
\begin{equation}
\label{eq:radar_2D_matrix}
\mathbf{A} = \begin{bmatrix}
\Delta \text{desc}_{0} & \Delta x_{0} & \Delta y_{0} & \Delta \theta_{0} \\
\Delta \text{desc}_{1} & \Delta x_{1} & \Delta y_{1} & \Delta \theta_{1} \\
\vdots & \vdots & \vdots & \vdots \\
\Delta \text{desc}_{n} & \Delta x_{n} & \Delta y_{n} & \Delta \theta_{n}
\end{bmatrix}
\end{equation}
The IMU encoder is based on a unidirectional Long Short-Term Memory (LSTM) network to handle accelerometer and gyroscope data, capturing temporal dependencies in the vehicle’s motion. The uni-directional nature of this LSTM network reflects the physics of the problem, as the future does not influence the current state in reality.\par

\subsection{Fusion Strategy}
\label{subsec:fusion}
\begin{figure}[htbp]
    \centering
    \includegraphics[width=0.4\textwidth]{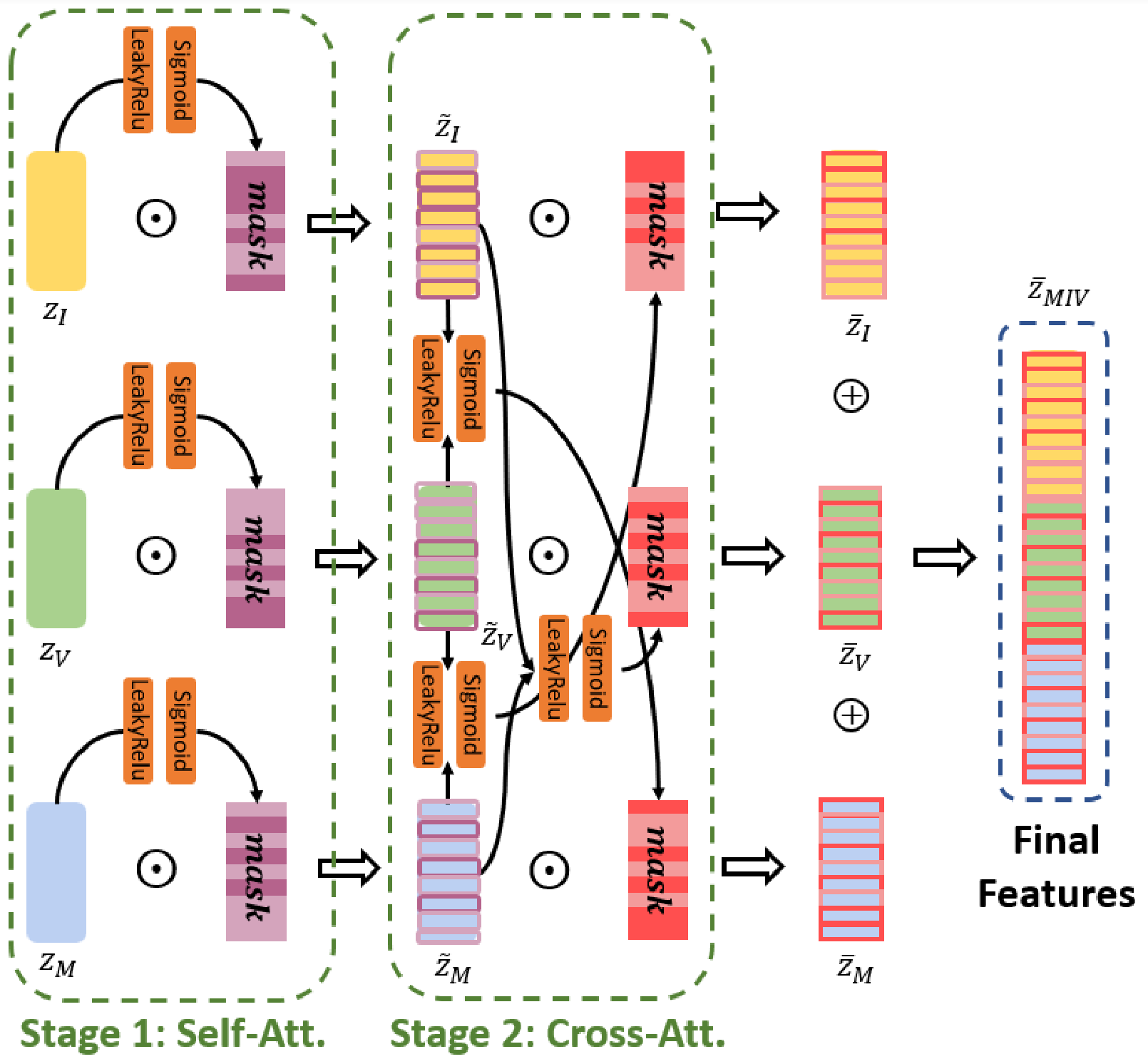}
    \caption{Mixed (two-stage) attention mechanism \cite{lu2020milliego}. The signs \( \oplus \) and \( \odot \) show vector appending and element-wise multiplication respectively.}
    \label{fig:two_stage_fusion}
\end{figure}

The outputs from the encoders are merged in the feature fusion module, shown in Fig.~\ref{fig:two_stage_fusion}. The module employs a 2-stage attention mechanism, self-attention and cross-attention, proposed by \cite{lu2020milliego}. The intuition behind self-attention stage is by letting individual odometry branches adapt themselves first and neglecting other sensors, similar to a self-filtering process that autonomously sieves informative features. This property is of fundamental importance for mmWave and optical odometry estimation, where widely separated spatial regions should jointly be considered. In this way, we can similarly generate self-regulated deep odometry features. As an instance of this stage computation shown in Eq.~\ref{eq:self_attention}, given an extracted feature vector $z_M$ by the radar encoder, a self-attention module first computes the similarity between two embedding spaces and uses the similarity to generate a mask $a_M$. Then the resulting features $\tilde{z}_M$ are given by applying the mask by Eq.~\ref{eq:self_attention},
\begin{equation}
    \begin{aligned}
        \tilde{z}_M &= a_M \odot z_M, \\
        \tilde{z}_V &= a_V \odot z_V, \\
        \tilde{z}_I &= a_I \odot z_I
    \end{aligned}
    \label{eq:self_attention}
\end{equation}
where $\odot$ denotes element-wise multiplication. Similarly, the other masked features computed in the self-attention stage are computed. The masks $a_M$, $a_V$, and $a_I$ are generated by passing the raw extracted features from 2 FC layers, applying Leaky Relu and sigmoid activation functions consecutively and producing values in the range [0,1] at the end. A value of 0 indicates that the extracted feature is completely ignored, while a value of 1 means the feature is fully utilized. These computations are shown in Eq.~\ref{eq:mask_computation},
\begin{equation}
    \begin{aligned}
        a_M &= \text{Sigmoid}(\mathcal{N}_{M,2}(\text{LeakyReLU}(\mathcal{N}_{M,1}(z_M)))), \\
        a_V &= \text{Sigmoid}(\mathcal{N}_{V,2}(\text{LeakyReLU}(\mathcal{N}_{V,1}(z_V)))), \\
        a_I &= \text{Sigmoid}(\mathcal{N}_{I,2}(\text{LeakyReLU}(\mathcal{N}_{I,1}(z_I))))
    \end{aligned}
    \label{eq:mask_computation}
\end{equation}
where \( \mathcal{N}_{i,j} \) denotes the \( j \)-th fully connected (FC) layer used for the \( i \)-th modality. Finally, these masked features enter the second fusion stage, which is the cross-attention stage.\par
The cross-attention serves as the second stage of reweighting, which examines the relationships between different modalities and is inspired by the mechanism of human perception. It is worth mentioning that the input features to the cross-attention computations are the masked features output from self-attention stage ($\tilde{z}_M$, $\tilde{z}_I$, and $\tilde{z}_C$). As an example of computations in this stage, the mask for radar derived in cross-attention stage $a_{VI \to M}$ is computed, and then it is multiplied elementwise into the masked radar features output from self-attention stage, as shown in Eq.~\ref{eq:cross_attention_radar} for radar features. The masks in this stage are computed in the same way as the self-attention masks while using outputs from the first stage instead of raw features.\par
\begin{equation}
    \begin{aligned}
        \bar{z}_M &= a_{VI \to M} \odot \tilde{z}_M, \\
        a_{VI \to M} &= \text{Sigmoid}(\mathcal{N}_{V,2}(\text{LeakyReLU}(\mathcal{N}_{V,1}(z_V \oplus z_I))))
    \end{aligned}
    \label{eq:cross_attention_radar}
\end{equation}
Finally, the resulting masked features from all modalities after the cross-attention stage are concatenated to form the final feature set, which is used as input for the subsequent stages of our sensor fusion model, as written in Eq.~\ref{eq:fused_features}:
\begin{equation}
    \bar{z}_{MIV} = \left[ a_{MV \to I} \odot \tilde{z}_I; \, a_{VI \to M} \odot \tilde{z}_M; \, a_{MI \to V} \odot \tilde{z}_V \right]
    \label{eq:fused_features}
\end{equation}

\subsection{Temporal Modeling and Task Solvers}
\label{subsec:temp_pose}
The fused features are integrated considering individual time stamps. This necessitates capturing temporal dependencies that exist between fused features from consecutive time stamps. The fused features are subsequently passed to an LSTM network, which captures the temporal dependencies and motion dynamics across consecutive frames. Finally, the output from the LSTM is fed into FC layers that map the temporal features to a 6-DoF motion estimate. The FC layers predict both the vehicle's translation and rotation, supervised by a 6-DoF loss function to ensure accurate motion estimation.

\section{Experiments and Results}
\label{experiments_results}

\subsection{Training Details}
\label{subsec:training_details}
We conducted experiments using the Boreas public dataset to evaluate the proposed multi-modal odometry estimation model. The model was trained on 30 scenes out of 44 available scenes, showcasing robustness across a variety of weather conditions, including snow, rain, sun glare, overcast, and cloudy scenarios. This diversity emphasizes the adaptability of the model in real-world environments. Our framework was implemented using PyTorch and trained on an NVIDIA GeForce RTX 4090. The training was conducted with a batch size of 8, using the Adam optimizer configured with a learning rate of $10^{-4}$, and run for 15 epochs.\\
We define an estimation frame as the time interval between two consecutive radar point clouds, including the associated camera images and IMU measurements recorded within that period. Since the number of camera images varies between estimation frames—typically ranging from two to five—we select only the two closest images to the radar timestamps. A similar strategy is applied to the IMU data: we uniformly sample 48 measurements per estimation frame to ensure a consistent input structure for the model.\\
The model's parameters are optimized using a Mean Squared Error (MSE) loss function. To account for varying scales in ground truth values across pose components, different weights are applied to balance the loss contributions for each variable. A mathematical formula (Eq.~\ref{eq:loss_weighting}) is proposed to compute the weight of every component based on the ground truth values mean and standard deviation, ensuring an equitable distribution of loss among all variables. The computed weights are shown in Table~\ref{tab:Computed_loss_weights}.
\begin{equation}
    w_i = \frac{1}{|\text{mean}(\text{pose}_i)| + \lambda \cdot \text{std}(\text{pose}_i)}
    \label{eq:loss_weighting}
\end{equation}

\begin{table}[H]
    \centering 
    \begin{tabular}{|p{4em} c|} 
    \hline
    \textbf{Pose} & \textbf{Computed weight} \\
    \hline \hline
    $\Delta{x}$ & 10.34 \\
    $\Delta{y}$ & 0.33 \\
    $\Delta{z}$ & 56.09 \\
    $\Delta{r}$ & 178.05 \\
    $\Delta{p}$ & 227.27 \\
    $\Delta{\gamma}$ & 39.05 \\
    \hline
    \end{tabular}
    \\[10pt]
    \caption{Computed weights for Different Pose Components}
    \label{tab:Computed_loss_weights}
\end{table}

\subsection{Results}
\label{subsec:results}
This section outlines the performance of the proposed model against the chosen baseline to estimate the ego-motion using the Boreas dataset. Regarding the evaluation metrics, we use the same metrics as the KITTI dataset \cite{geiger2013vision} suggested by Boreas. The KITTI odometry metrics average the relative position and orientation errors over every length subsequence (100m, 200m, 300m,..., 800m). This results in two metrics, a translational drift reported as a
percentage of path length, and a rotational drift reported as
degrees per meter traveled. To establish a baseline for comparison, we implemented an alternative model using the same architecture as our proposed model but replaced the 2-stage attention mechanism with a soft fusion approach proposed by Chen et al. \cite{chen2019selective}.\par
The translation error of the two models, reported as a percentage of the total distance traveled, and the rotation error, expressed in degrees per 100 meters, are summarized in Table~\ref{tab:length_based_results} based on length. The results indicate that while the baseline model achieves a lower average translation error (5.06\%) compared to Raci-Net (5.63\%), our proposed model consistently demonstrates superior performance in rotation accuracy, with an average rotation error of 1.13 \text{$^{\circ}$}/100\text{m} versus 1.35 \text{$^{\circ}$}/100\text{m} for the baseline. This highlights that while the overall translation performance of Raci-Net may not exceed the baseline, it excels in maintaining orientation accuracy, which is crucial for tasks that require precise rotational alignment.\par
\begin{table}[htbp]
    \centering
    \caption{Translation and Rotation Errors in different sub-lengths for our proposed model and the baseline \cite{chen2019selective}. $t_{err}$: average translational RMSE drift \(\%\) on length of 100 m–800 m. $r_{err}$: average rotational RMSE drift ( ° / 100 m) on length of 100m–800m. The values are averaged over all scenes to have only the effect of lengths.}
    \label{tab:length_based_results}
    \begin{tabular}{lcccc}
        \hline
        Segment Length & \textbf{$t_{\text{err}, \text{ours}}$ (\%)} & \textbf{$t_{\text{err}, \text{base}}$ (\%)} & \textbf{$r_{\text{err}, \text{ours}}$ } & \textbf{$r_{\text{err}, \text{base}}$ } \\
        \hline
        100m & 6.19 & 5.41 & 2.64 & 1.82 \\
        200m & 6.10 & 4.76 & 1.36 & 1.49 \\
        300m & 5.92 & 4.63 & 1.00 & 1.37 \\
        400m & 5.36 & 4.72 & 0.92 & 1.31 \\
        500m & 5.71 & 4.65 & 0.89 & 1.24 \\
        600m & 5.65 & 5.14 & 0.84 & 1.23 \\
        700m & 5.02 & 5.31 & 0.68 & 1.16 \\
        800m & 5.09 & 5.85 & 0.70 & 1.17 \\
        \hline
        Avg & 5.63 & 5.06 & 1.13 & 1.35 \\
        \hline
    \end{tabular}
\end{table}

\begin{figure}[H]
    \centering
    \includegraphics[width=0.4\textwidth]{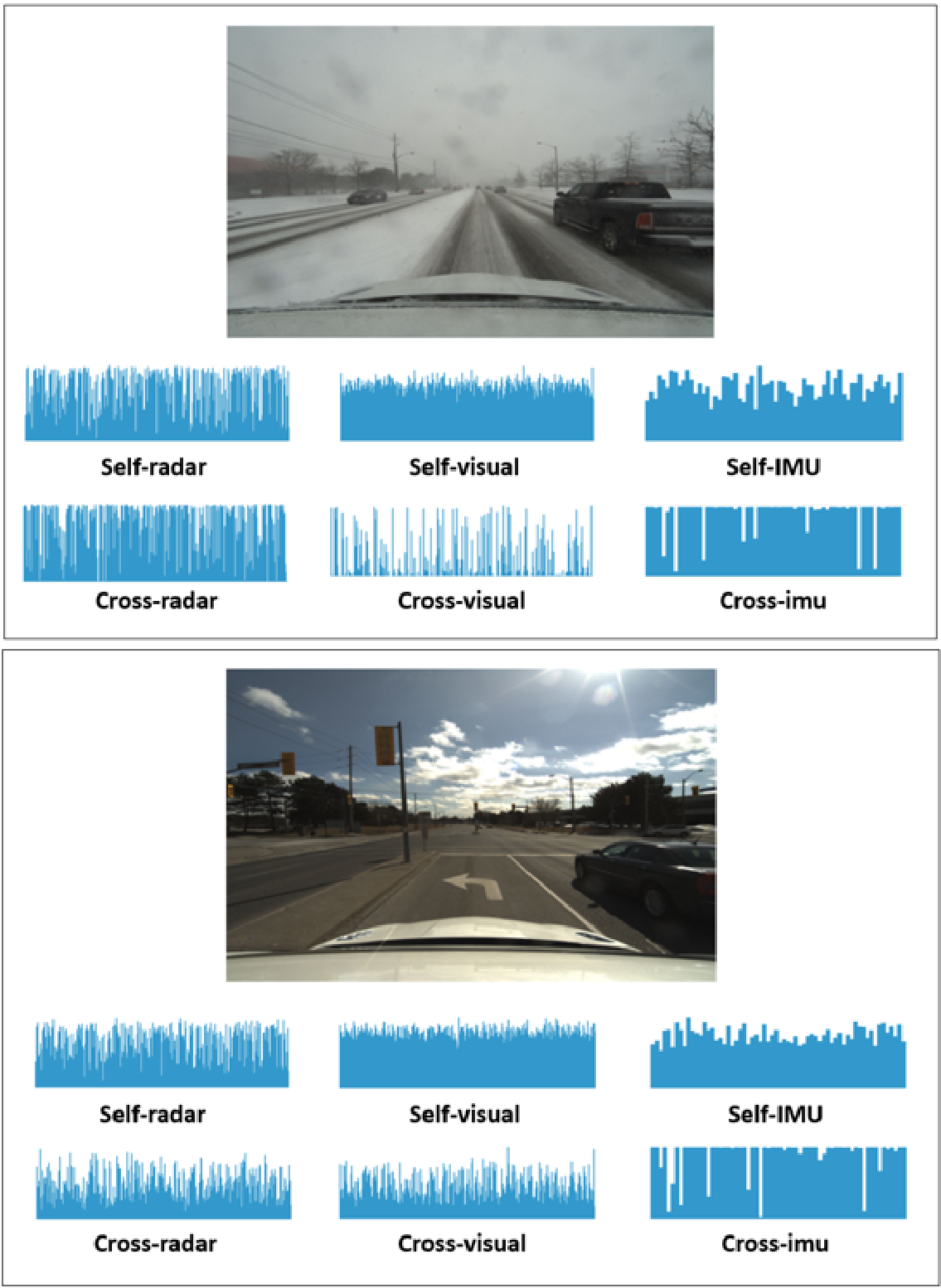}
    \caption{Computed Masks in a Snowy and Sunny Scenarios}
    \label{fig:Masks_Snowing_and_Sunny_Test}
\end{figure}

Table~\ref{tab:scene_errors} provides the same results in a weather-based manner, showing the resilience of the model in varying environmental scenarios, such as snow, rain, overcast, cloudy, and sunny, with consistent performance under varying conditions.\\
Attention masks generated by the FC layers in the fusion block play a crucial role in feature weighting. As an example, Fig.~\ref{fig:Masks_Snowing_and_Sunny_Test} shows masks in snowy and sunny scenarios, showing adaptive weighting where more reliable sensor data are prioritized. In these figures, the horizontal axis represents the weight elements within the computed mask, while the vertical axis indicates the intensity of the weights, which range from 0 to 1. As depicted, in snowy conditions, poor visibility and snow accumulation on the lens lead to lower attention weights for camera features, while IMU and mmWave radar receive higher weights due to their reliability. Conversely, in sunny scenario, all the modalities are reliable, and the model assigns balanced attention across the sensors.\\
\begin{table}[htbp]
    \centering
    \small
    \caption{Translation and Rotation Errors in different weather conditions from the proposed and the baseline model \cite{chen2019selective}. The values are averaged over all sub-lengths to have only the effect of weather.}
    \label{tab:scene_errors}
    \begin{tabular}{p{8em} cccc}
    \hline
    \textbf{Scene} & \textbf{$t_{\text{err}, \text{ours}}$ } & \textbf{$t_{\text{err}, \text{base}}$} & \textbf{$r_{\text{err}, \text{ours}}$ } & \textbf{$r_{\text{err}, \text{base}}$ } \\
    \hline
    2020-11-26-13-58 overcast & 5.82 & 4.43 & 0.74 & 1.11 \\
    2021-01-19-15-08 cloudy & 4.71 & 5.75 & 0.73 & 0.88 \\
    2021-01-26-11-22 Snowing & 5.85 & 6.12 & 0.77 & 1.45 \\
    2021-04-29-15-55 rainy & 6.98 & 5.95 & 0.84 & 1.70 \\
    2021-06-03-16-00 sunny & 4.30 & 5.27 & 0.75 & 1.41 \\
    \hline
    \end{tabular}
\end{table}

\section{Discussion}
\label{sec:discussion}
Regarding the result reported in Table~\ref{tab:length_based_results}, the trend indicates that while translation errors are present, they don't significantly increase with distance, suggesting that the model effectively maintains its performance even over long ranges. It is crucial to recognize that the downward trend in reported translation errors does not imply the absence of cumulative drift; The downward trend in translation errors as segment lengths increase suggests that the model benefits from averaging out positional deviations over longer distances. While shorter segments may show higher relative errors due to abrupt or localized inaccuracies, these errors tend to be smoothed out over longer trajectories. This implies that the model’s performance stabilizes as the cumulative effect of smaller deviations is distributed across an extended path, reducing the proportional impact of individual errors. This trend indicates that the model is more reliable in maintaining consistent odometry estimation when dealing with longer trajectories, demonstrating its capacity to effectively manage drift and maintain stable performance over extended driving distances. Similarly, the reduction in average rotation error is due to the averaging effect over longer segments; while sharp turns contribute more significantly to higher rotation errors in shorter segments, their impact becomes less pronounced when averaged out over longer distances.\par

\section{Conclusion and Future Work}
\label{sec:Conclusion}
This paper explored a multi-modal ego-motion estimation system to enhance autonomous vehicle odometry estimation under challenging adverse weather conditions. By fusing mmWave radar, visual, and inertial data within a deep learning framework, robust 6-DoF pose estimation was achieved.\par
The proposed model outperformed the baseline in rotation estimation and long-range translation estimation, highlighting the proposed radar encoder's success in contributing to the odometry estimation task and its ability to extract meaningful features, as well as the benefit of a more structured attention mechanism for improved localization.\par
Potential areas for improvement include:
\begin{itemize}
    \item \textbf{Motion Estimation Frequency:} Implementing a method to increase the radar frequency which is the frequency of motion estimation.
    \item \textbf{3D Sensor Data Integration:} Using sensors like LiDAR for richer spatial information.`
    \item \textbf{Environmental Adaptability:} Further adjustment to the weights to cover a wider array of challenges such as common technical failures
    \item \textbf{GNSS Integration:} Incorporating GNSS data can close the loop and reduce drift by providing global pose information, enhancing odometry accuracy.
\end{itemize}
In conclusion, this work advanced autonomous vehicle localization by proposing a robust radar encoder and leveraging multi-modal fusion with attention-based mechanisms. The findings support future research into reliable navigation systems that can perform across diverse weather conditions.\\

\bibliographystyle{./bibliography/IEEEtran}
\bibliography{./bibliography/IEEEabrv}

\begin{thebibliography}{10}
\providecommand{\url}[1]{#1}
\csname url@samestyle\endcsname
\providecommand{\newblock}{\relax}
\providecommand{\bibinfo}[2]{#2}
\providecommand{\BIBentrySTDinterwordspacing}{\spaceskip=0pt\relax}
\providecommand{\BIBentryALTinterwordstretchfactor}{4}
\providecommand{\BIBentryALTinterwordspacing}{\spaceskip=\fontdimen2\font plus
\BIBentryALTinterwordstretchfactor\fontdimen3\font minus \fontdimen4\font\relax}
\providecommand{\BIBforeignlanguage}[2]{{%
\expandafter\ifx\csname l@#1\endcsname\relax
\typeout{** WARNING: IEEEtran.bst: No hyphenation pattern has been}%
\typeout{** loaded for the language `#1'. Using the pattern for}%
\typeout{** the default language instead.}%
\else
\language=\csname l@#1\endcsname
\fi
#2}}
\providecommand{\BIBdecl}{\relax}
\BIBdecl

\bibitem{dahal2023vehicle}
P.~Dahal, J.~Prakash, S.~Arrigoni, and F.~Braghin, ``Vehicle state estimation through modular factor graph-based fusion of multiple sensors,'' 2023.

\bibitem{RobustStateNet}
\BIBentryALTinterwordspacing
P.~Dahal, S.~Mentasti, L.~Paparusso, S.~Arrigoni, and F.~Braghin, ``Robuststatenet: Robust ego vehicle state estimation for autonomous driving,'' \emph{Robotics and Autonomous Systems}, p. 104585, 2023. [Online]. Available: \url{https://www.sciencedirect.com/science/article/pii/S0921889023002245}
\BIBentrySTDinterwordspacing

\bibitem{msf}
J.~Nubert, S.~Khattak, and M.~Hutter, ``Graph-based multi-sensor fusion for consistent localization of autonomous construction robots,'' in \emph{2022 IEEE International Conference on Robotics and Automation (ICRA)}.\hskip 1em plus 0.5em minus 0.4em\relax IEEE, 2022.

\bibitem{Integrated_Ego_Estimation}
M.~Bersani, S.~Mentasti, P.~Dahal, S.~Arrigoni, M.~Vignati, F.~Cheli, and M.~Matteucci, ``An integrated algorithm for ego-vehicle and obstacles state estimation for autonomous driving,'' \emph{Robotics and Autonomous Systems}, vol. 139, p. 103662, 2021.

\bibitem{Millimeter_Wave_Human_Blockage}
G.~R. MacCartney, S.~Deng, S.~Sun, and T.~S. Rappaport, ``Millimeter-wave human blockage at 73 ghz with a simple double knife-edge diffraction model and extension for directional antennas,'' in \emph{2016 IEEE 84th Vehicular Technology Conference (VTC-Fall)}.\hskip 1em plus 0.5em minus 0.4em\relax IEEE, 2016.

\bibitem{lu2020milliego}
\BIBentryALTinterwordspacing
C.~X. Lu, M.~R.~U. Saputra, P.~Zhao, Y.~Almalioglu, P.~P.~B. de~Gusmao, C.~Chen, K.~Sun, N.~Trigoni, and A.~Markham, ``milliego: Single-chip mmwave radar aided egomotion estimation via deep sensor fusion,'' in \emph{Proceedings of the 18th ACM Conference on Embedded Networked Sensor Systems (SenSys)}.\hskip 1em plus 0.5em minus 0.4em\relax Virtual Event, Japan: ACM, November 2020, pp. 6433--6440, available from ACM Digital Library. [Online]. Available: \url{https://doi.org/10.1145/3384419.3430776}
\BIBentrySTDinterwordspacing

\bibitem{vargas2021overview}
\BIBentryALTinterwordspacing
J.~Vargas, S.~Alsweiss, O.~Toker, R.~Razdan, and J.~Santos, ``An overview of autonomous vehicles sensors and their vulnerability to weather conditions,'' \emph{Sensors}, vol.~21, p. 5397, August 2021, available under the Creative Commons Attribution (CC BY) license. [Online]. Available: \url{https://doi.org/10.3390/s21165397}
\BIBentrySTDinterwordspacing

\bibitem{ORB_SLAM}
T.~S. . D.~C. Jakob~Engel, ``Lsd-slam: Large-scale direct monocular slam,'' in \emph{Computer Vision – ECCV 2014}.\hskip 1em plus 0.5em minus 0.4em\relax Springer, 2015.

\bibitem{LSD_SLAM}
J.~D.~T. Raul Mur-Artal, J. M. M.~Montiel, ``Orb-slam: a versatile and accurate monocular slam system,'' in \emph{IEEE Transactions on Robotics}.\hskip 1em plus 0.5em minus 0.4em\relax IEEE, 2015.

\bibitem{lim2020review}
K.~L. Lim and T.~Bräunl, ``A review of visual odometry methods and its applications for autonomous driving,'' \emph{arXiv preprint arXiv:2009.09193v1}, September 2020, available at: https://arxiv.org/abs/2009.09193.

\bibitem{cheng2022review}
\BIBentryALTinterwordspacing
J.~Cheng, L.~Zhang, Q.~Chen, X.~Hu, and J.~Cai, ``A review of visual slam methods for autonomous driving vehicles,'' \emph{Engineering Applications of Artificial Intelligence}, vol. 114, p. 104992, June 2022, received 9 October 2021; Revised 12 May 2022; Accepted 19 May 2022. [Online]. Available: \url{https://doi.org/10.1016/j.engappai.2022.104992}
\BIBentrySTDinterwordspacing

\bibitem{bala2022advances}
\BIBentryALTinterwordspacing
J.~A. Bala, S.~A. Adeshina, and A.~M. Aibinu, ``Advances in visual simultaneous localisation and mapping techniques for autonomous vehicles: A review,'' \emph{Sensors}, vol.~22, p. 8943, November 2022, available under the Creative Commons Attribution (CC BY) license. [Online]. Available: \url{https://doi.org/10.3390/s22228943}
\BIBentrySTDinterwordspacing

\bibitem{tang2023comparative}
\BIBentryALTinterwordspacing
Q.~Tang, J.~Liang, and F.~Zhu, ``A comparative review on multi-modal sensors fusion based on deep learning,'' \emph{Signal Processing}, vol. 213, p. 109165, July 2023, available online 3 July 2023. [Online]. Available: \url{https://doi.org/10.1016/j.sigpro.2023.109165}
\BIBentrySTDinterwordspacing

\bibitem{wang2017deepvo}
\BIBentryALTinterwordspacing
S.~Wang, R.~Clark, H.~Wen, and N.~Trigoni, ``Deepvo: Towards end-to-end visual odometry with deep recurrent convolutional neural networks,'' in \emph{Proceedings of the IEEE International Conference on Robotics and Automation (ICRA)}.\hskip 1em plus 0.5em minus 0.4em\relax Singapore: IEEE, May 2017, pp. 2043--2050. [Online]. Available: \url{https://doi.org/10.1109/ICRA.2017.7989236}
\BIBentrySTDinterwordspacing

\bibitem{clark2017vinet}
R.~Clark, S.~Wang, H.~Wen, A.~Markham, and N.~Trigoni, ``Vinet: Visual-inertial odometry as a sequence-to-sequence learning problem,'' in \emph{Proceedings of the Thirty-First AAAI Conference on Artificial Intelligence}.\hskip 1em plus 0.5em minus 0.4em\relax San Francisco, CA, USA: AAAI Press, February 2017, pp. 3995--4001, available from AAAI Digital Library.

\bibitem{kendall2015posenet}
\BIBentryALTinterwordspacing
A.~Kendall, M.~Grimes, and R.~Cipolla, ``Posenet: A convolutional network for real-time 6-dof camera relocalization,'' in \emph{Proceedings of the IEEE International Conference on Computer Vision (ICCV)}.\hskip 1em plus 0.5em minus 0.4em\relax IEEE, December 2015, pp. 2938--2946, presented at ICCV 2015, Santiago, Chile. [Online]. Available: \url{https://doi.org/10.1109/ICCV.2015.336}
\BIBentrySTDinterwordspacing

\bibitem{brahmbhatt2018geometry}
\BIBentryALTinterwordspacing
S.~Brahmbhatt, J.~Gu, K.~Kim, J.~Hays, and J.~Kautz, ``Geometry-aware learning of maps for camera localization,'' in \emph{Proceedings of the IEEE Conference on Computer Vision and Pattern Recognition (CVPR)}.\hskip 1em plus 0.5em minus 0.4em\relax Salt Lake City, UT, USA: IEEE, June 2018, pp. 2616--2625, available from IEEE Xplore. [Online]. Available: \url{https://doi.org/10.1109/CVPR.2018.00276}
\BIBentrySTDinterwordspacing

\bibitem{clark2017vidloc}
\BIBentryALTinterwordspacing
R.~Clark, S.~Wang, A.~Markham, N.~Trigoni, and H.~Wen, ``Vidloc: A deep spatio-temporal model for 6-dof video-clip relocalization,'' in \emph{Proceedings of the IEEE Conference on Computer Vision and Pattern Recognition (CVPR)}.\hskip 1em plus 0.5em minus 0.4em\relax Honolulu, HI, USA: IEEE, July 2017, available from IEEE Xplore. [Online]. Available: \url{https://doi.org/10.1109/CVPR.2017.702}
\BIBentrySTDinterwordspacing

\bibitem{chen2019selective}
\BIBentryALTinterwordspacing
C.~Chen, S.~Rosa, Y.~Miao, C.~X. Lu, W.~Wu, A.~Markham, and N.~Trigoni, ``Selective sensor fusion for neural visual-inertial odometry,'' in \emph{Proceedings of the IEEE/CVF Conference on Computer Vision and Pattern Recognition (CVPR)}.\hskip 1em plus 0.5em minus 0.4em\relax Long Beach, CA, USA: IEEE, June 2019, pp. 10\,534--10\,543, presented at CVPR 2019. [Online]. Available: \url{https://doi.org/10.1109/CVPR.2019.01079}
\BIBentrySTDinterwordspacing

\bibitem{dahal2023fault}
\BIBentryALTinterwordspacing
P.~Dahal, S.~Mentasti, L.~Paparusso, S.~Arrigoni, and F.~Braghin, ``Fault resistant odometry estimation using message passing neural network,'' in \emph{Proceedings of the IEEE Intelligent Vehicles Symposium (IV)}.\hskip 1em plus 0.5em minus 0.4em\relax Milano, Italy: IEEE, June 2023, pp. 978--986, available from IEEE Xplore. [Online]. Available: \url{https://doi.org/10.1109/IV55152.2023.10186649}
\BIBentrySTDinterwordspacing

\bibitem{saputra2020deeptio}
\BIBentryALTinterwordspacing
M.~R.~U. Saputra, P.~P.~B. de~Gusmao, C.~X. Lu, Y.~Almalioglu, S.~Rosa, C.~Chen, J.~Wahlstrom, W.~Wang, A.~Markham, and N.~Trigoni, ``Deeptio: A deep thermal-inertial odometry with visual hallucination,'' \emph{arXiv preprint arXiv:1909.07231v2}, January 2020, available on arXiv. [Online]. Available: \url{https://arxiv.org/abs/1909.07231v2}
\BIBentrySTDinterwordspacing

\bibitem{richardson2011strengths}
\BIBentryALTinterwordspacing
J.~J. Richardson and L.~M. Moskal, ``Strengths and limitations of assessing forest density and spatial configuration with aerial lidar,'' \emph{Remote Sensing of Environment}, vol. 115, pp. 2640--2651, June 2011, available from ScienceDirect. [Online]. Available: \url{https://doi.org/10.1016/j.rse.2011.05.020}
\BIBentrySTDinterwordspacing

\bibitem{cen2018precise}
\BIBentryALTinterwordspacing
S.~H. Cen and P.~Newman, ``Precise ego-motion estimation with millimeter-wave radar under diverse and challenging conditions,'' in \emph{Proceedings of the IEEE International Conference on Robotics and Automation (ICRA)}.\hskip 1em plus 0.5em minus 0.4em\relax Brisbane, Australia: IEEE, May 2018, pp. 6045--6052, available from IEEE Xplore. [Online]. Available: \url{https://doi.org/10.1109/ICRA.2018.8462925}
\BIBentrySTDinterwordspacing

\bibitem{kung2023ndt}
P.-C. Kung, C.-C. Wang, and W.-C. Lin, ``A normal distribution transform-based radar odometry designed for scanning and automotive radars,'' in \emph{Proceedings of the IEEE International Conference on Robotics and Automation (ICRA)}.\hskip 1em plus 0.5em minus 0.4em\relax London, UK: IEEE, March 2023, pp. 6045--6052, available from IEEE Xplore.

\bibitem{burnett2021motion}
\BIBentryALTinterwordspacing
K.~Burnett, A.~P. Schoellig, and T.~D. Barfoot, ``Do we need to compensate for motion distortion and doppler effects in spinning radar navigation?'' \emph{IEEE Robotics and Automation Letters}, January 2021, preprint version, accepted January 2021. [Online]. Available: \url{https://github.com/keenan-burnett/yeti\_radar\_odometry}
\BIBentrySTDinterwordspacing

\bibitem{barnes2020under}
D.~Barnes and I.~Posner, ``Under the radar: Learning to predict robust keypoints for odometry estimation and metric localisation in radar,'' in \emph{Proceedings of the IEEE International Conference on Robotics and Automation (ICRA)}.\hskip 1em plus 0.5em minus 0.4em\relax Paris, France: IEEE, May 2020, presented at ICRA 2020, available from IEEE Xplore.

\bibitem{teed2020raft}
\BIBentryALTinterwordspacing
Z.~Teed and J.~Deng, ``Raft: Recurrent all-pairs field transforms for optical flow,'' in \emph{Proceedings of the IEEE/CVF Conference on Computer Vision and Pattern Recognition (CVPR)}.\hskip 1em plus 0.5em minus 0.4em\relax Seattle, WA, USA: IEEE, June 2020, pp. 6778--6788, available from IEEE Xplore. [Online]. Available: \url{https://doi.org/10.1109/CVPR42600.2020.00681}
\BIBentrySTDinterwordspacing

\bibitem{geiger2013vision}
\BIBentryALTinterwordspacing
A.~Geiger, P.~Lenz, C.~Stiller, and R.~Urtasun, ``Vision meets robotics: The kitti dataset,'' \emph{International Journal of Robotics Research}, vol.~32, no.~11, pp. 1231--1237, 2013, available from KITTI Vision Benchmark Suite. [Online]. Available: \url{https://doi.org/10.1177/0278364913491297}
\BIBentrySTDinterwordspacing

\end{thebibliography}

\end{document}